\documentclass[submission,copyright,creativecommons]{eptcs}
\providecommand{\event}{ICLP 2020} 
\usepackage{underscore}           

\usepackage[ruled,vlined]{algorithm2e}
\usepackage{graphicx}
\usepackage{float}

\usepackage{times}
\usepackage{helvet}
\usepackage{courier}
\usepackage{xspace}

\usepackage{todonotes}

\def\clingo{{\sc Clingo}\xspace}

\def\ba{\begin{array}}
\def\ea{\end{array}}
\def\beq{\begin{equation}}
\def\eeq#1{\label{#1}\end{equation}}

\def\gM{{\bf M}\xspace}
\def\gC{{\bf C}\xspace}
\def\gD{{\bf D}\xspace}
\def\gN{{\bf N}\xspace}
\def\gE{{\bf E}\xspace}

\title{Solving Gossip Problems using Answer Set
Programming: \\ An Epistemic Planning Approach}
\author{Esra Erdem
\institute{Sabanci University, Faculty of Engineering and Natural Sciences, Istanbul, Turkey}
\email{esraerdem@sabanciuniv.edu}
\and
Andreas Herzig
\institute{Institut de Recherche en Informatique de Toulouse, Universit\'{e} Paul Sabatier, Toulouse, France}
\email{Andreas.Herzig@irit.fr }
}

\def\titlerunning{Gossip Problems using Answer Set Programming}
\def\authorrunning{Erdem and Herzig}

\begin{document}

\maketitle

\begin{abstract}
We investigate the use of Answer Set Programming to solve variations
of gossip problems, by modeling them as epistemic planning problems.
\end{abstract}

\section{Introduction}

The gossip problem is described by Hedetniemi et al. in their survey~\cite{HedetHL88} as follows:

\begin{quote}
Gossiping refers to the information dissemination problem that exists when each
member of a set $A$ of $n$ individuals knows a unique piece of information and must
transmit it to every other person. The problem is solved by producing a sequence of
unordered pairs $(i,j)$, $i,j \in A$, each of which represents a phone call made between a
pair of individuals, such that during each call the two people involved exchange all
of the information they know at that time; and such that at the end of the sequence
of calls, everybody knows everything. Such a calling sequence, which completes
gossiping among the $n$ people, is called {\em complete}.
\end{quote}

The gossip problem has been studied by many researchers, in particular, in the context of communication networks. While the most widely studied variant is the following optimization problem:

\begin{itemize}
\item[\gM] Minimize the number of calls in a complete calling sequence.
\end{itemize}

\noindent
there are other interesting variants considering the communication network and the sharing of gossips:

\begin{itemize}
\item[\gC] Parallel communications: Concurrency of calls is allowed, but not two calls to and/or from the same agent at the same time.
\item[\gD] Directional gossips: Communication is limited from some agents to some other agents; it does not have to be bi-directional.
\item[\gN] Negative goals: Some agents should not know the secrets of some other agents in the end.
\end{itemize}

\noindent The variations \gD\ and \gN\ of the gossip problem are intractable~\cite[Section~4]{Krumme92gossipingin} \cite[Theorem~10]{CooperHMMR16arxiv}. Optimal protocols for these variants are studied in \cite{CooperHMMR19}.

We are also interested in the following variation of the gossip problem, from the perspective of multiagent epistemic reasoning:

\begin{itemize}
\item[\gE] Higher-order epistemic goals: Some agents should know that some other specified agents know that ... that some other agents know some secrets in the end.
\end{itemize}

Other variants exist, for example dynamic gossip where the communication graph evolves, which can be thought of as agents communicating telephone numbers of other agents during a call \cite{DitmarschEPRS17}; we donot study these dynamic variants in this paper.

We consider the mathematical definitions of the gossip problems as in~\cite{CooperHMMR16arxiv}, and illustrate how they can be solved using Answer Set Programming  (ASP)~\cite{BrewkaET11} in the spirit of epistemic planning~\cite{CooperHMMR16} where the gossip problem and its variants can be viewed as a paradigmatic problem that provides several benchmarks. Secrets are viewed as propositions that are either true or false.

\section{Representing and Solving Gossip Problems in ASP}

We represent the gossip problem and its variations as epistemic planning problems in the ASP language ASP-Core-2~\cite{asp2core}, and utilize the ASP solver~\clingo~\cite{GebserKKOSS11} to compute solutions.

In ASP-Core-2, the object/predicate constants are denoted by strings that start with lower-case letters, and variables are denoted by strings that start with upper-case letters, very much like in Prolog.

\subsection{Domain predicates}

The atemporal input of the gossip problem are represented by atoms of the forms

\begin{itemize}
\item {\small\tt agent(i)} -- {\small\tt i} denotes an agent,
\item {\small\tt secret(x)} -- {\small\tt x} denotes a secret that may be known and shared by agents,
\item {\small\tt connected(i,j)} -- agent {\small\tt i} can pass a secret to agent {\small\tt j}.
\end{itemize}

\noindent  These atoms are used to specify the domains of variables used in the formulas, and thus they are called ``domain predicates''.

If there are {\small\tt n} agents and {\small\tt s} secrets then these atoms can be represented in a more compact as follows:

{\small \begin{verbatim}
agent(1..n).
secret(1..s).
\end{verbatim}}

\noindent For a complete communication network, where every agent can pass their secrets to every other agent, {\small\tt connected(i,j)} can be defined as follows:

{\small \begin{verbatim}
connected(I,J) :- agent(I), agent(J), I!=J.
\end{verbatim}}

\noindent For variation \gD\ of the gossip problem, if the communication network is not complete, {\small\tt connected(i,j)} can be defined differently.

\subsection{Auxiliary definitions: Pieces of information}

Secrets and pieces of information that the agents know and can communicate to each other are defined recursively with respect to an epistemic depth.

The epistemic depth is specified by atoms of the form {\small\tt depth(y)} as follows:

{\small \begin{verbatim}
depth(0..d).
\end{verbatim}}

\noindent where the constant {\small\tt d} specifies the maximum depth.

As the base case, at epistemic depth 0, every secret {\small\tt K} can be a piece of information that an agent knows (i.e., the agent knows-whether~{\small\tt K}).

{\small \begin{verbatim}
info(K,0) :- secret(K).
\end{verbatim}}

\noindent Intuitively, {\small\tt info(K,0)} expresses that {\small\tt K} is a piece of information of depth 0.

At epistemic depth 1, we consider pieces of information of the form ``an agent {\small\tt I} knows-whether {\small\tt K}''.

{\small \begin{verbatim}
info(kw(I,K),1) :- agent(I), info(K,0).
\end{verbatim}}

Here {\small\tt kw(I,K)} corresponds to ${\bf Kw}_I K$ of DL-PA~\cite{BalbianiHT13} as adapted by Cooper et al. \cite{CooperHMMR16}.

At an epistemic depth {\small\tt D} greater than 1, pieces of information are nested like ``an agent {\small\tt I} knows-whether agent {\small\tt J} knows-whether {\small\tt K}'', and can be defined as follows:

{\small \begin{verbatim}
info(kw(I,kw(J,K)),D) :- agent(I;J), I!=J, info(kw(J,K),D-1), depth(D), D>1.
\end{verbatim}}

Since we are interested in solving gossip problems as planning problems, introspective statements are, in a sense, irrelevant; this is why {\small\tt I!=J} in the body of the rule above.

\subsection{Fluents: Knowledge of agents}

The temporal input/output/constraints of the gossip problem (e.g., which agent knows-whether or should know-whether which secret, initially, at some time step, or in the end) can be represented with fluent constants -- atoms of the form

\begin{itemize}
\item[]
{\small\tt kww(i,x,t)} --  agent {\small\tt i} knows-whether secret {\small\tt x} at step~{\small\tt t}.
\end{itemize}

Time steps, starting from {\small\tt 0} to a given upper bound {\small\tt m} on makespan of plans, are specified as follows:

{\small \begin{verbatim}
time(0..m).
\end{verbatim}}

\paragraph{Initial values of fluents} Initially, the agents may already know-whether some secrets {\small\tt K} (at epistemic depth~{\small\tt 0}):

{\small \begin{verbatim}
{kww(I,info(K,0),0)} :- agent(I), info(K,0).
\end{verbatim}}

\noindent as well as some pieces of information about themselves or other agents (at epistemic depth {\small\tt D}):

{\small \begin{verbatim}
{kww(I,info(kw(J,K),D),0)} :- agent(I), I!=J, info(kw(J,K),D).
\end{verbatim}}


Further conditions can be presented about the initial state by means of constraints, as illustrated by the following examples.

\begin{itemize}
\item  Initially, every agent {\small\tt I} knows-whether secret {\small\tt I}:

{\small \begin{verbatim}
:- not kww(I,info(I,0),0), agent(I), info(I,0).
\end{verbatim}}

\noindent and they do not know any other secrets, or pieces of information of epistemic depth greater than 0:

{\small \begin{verbatim}
:- kww(I,info(K,0),0), agent(I), info(K,0), I!=K.
:- kww(I,info(K,D),0), agent(I), info(K,D), D>0.
\end{verbatim}}

\noindent Cooper et al.~(\cite[Example~2]{CooperHMMR16}) consider the initial state of the gossip problem, as in the example above: $s_0 = \{ {\bf Kw}_i s_i : 1\leq i\leq n\}$.

\item Initially, Agent 1 knows-whether Agent 2 knows-whether secret 2:

{\small \begin{verbatim}
:- not kww(1,info(kw(2,2),1),0).
\end{verbatim}}

\item Initially, every agent knows-whether at least {\small\tt l} secret:

{\small \begin{verbatim}
:- not l {kww(I,info(K,0),0) : info(K,0)}, agent(I).
\end{verbatim}}

\noindent and/or at most {\small\tt u} secrets:

{\small \begin{verbatim}
:- {kww(I,info(K,0),0) : info(K,0)} u, agent(I).
\end{verbatim}}

\item Initially, no two agents know-whether the same piece of information:

{\small \begin{verbatim}
:- 2 {kww(I,info(K,_),0): agent(I)}, info(K,_).
\end{verbatim}}

\end{itemize}

\paragraph{Goal values of fluents}
Goal conditions can be represented by constraints as well, as illustrated in the following examples.

\begin{itemize}
\item Every agent knows-whether at least {\small\tt N} pieces of information:

{\small \begin{verbatim}
goal(T) :-
  N {kww(I,info(K,0),T): agent(I), info(K,0) ;
     kww(I,info(K,D),T): agent(I), info(K,D), depth(D), D>0},
  infoNo(N), time(T).
\end{verbatim}}

The secrets are of epistemic depth {\small\tt 0} (i.e., {\small\tt info(K,0)}), whereas
other pieces of information are of larger epistemic depths (i.e., {\small\tt info(K,D)} where {\small\tt D>0}).

The number {\small\tt N} can be specified as a constant, e.g., by a fact, as follows:

{\small \begin{verbatim}
infoNo(2).
\end{verbatim}}

Alternatively, it can be defined as the total number of pieces of information of maximum epistemic depth {\small\tt d}, for  introspective {\small\tt n} agents and {\small\tt s} unique secrets, as follows:

{\small \begin{verbatim}
infoNo(N) :- infoNoAux(N,_,d).
infoNoAux(s,s,0).
infoNoAux(n*N+N1,n*N,D+1) :- infoNoAux(N1,N,D), depth(D), depth(D+1).
\end{verbatim}}

Here, auxiliary atoms of the form {\small\tt infoNoAux(N1,N,D)} represent the total number {\small\tt N1} of pieces of information of maximum epistemic depth {\small\tt D}, provided that the total number of pieces of information of epistemic depth {\small\tt D-1} is {\small\tt N}. The second line above expresses that, the number of secrets that might be known by the agents is {\small\tt s}. The third line defines the total number of pieces of information over epistemic depths {\small\tt D>0} recursively. If the number of pieces of information of epistemic depth {\small\tt D} is {\small\tt N}, then the maximum number of pieces of information of epistemic depth {\small\tt D+1} is {\small\tt n*N}. Therefore, if the total number of pieces of information of maximum epistemic depth {\small\tt D} is {\small\tt N1}, then the total number of pieces of information of maximum epistemic depth {\small\tt D+1} is {\small\tt n*N+N1}.

\item Some agents should know-whether some secrets but not know-whether some other secrets. For instance, the following goal conditions express that agent {\small\tt 1} should know-whether secrets {\small\tt 1} and  {\small\tt 3} but not secret {\small\tt 2}:

{\small \begin{verbatim}
goal(1,T) :- kww(1,info(1,0),T), info(1,0),
  kww(1,info(3,0),T), info(3,0),
  not kww(1,info(2,0),T), info(2,0), time(T).
\end{verbatim}}

whereas other agents {\small\tt I} know-whether all~{\small\tt s} messages:

{\small \begin{verbatim}
goal(I,T) :- agent(I), I!=1,
  s {kww(I,info(K,0),T): info(K,0)}, time(T).
\end{verbatim}}

The goal is reached when these goal conditions hold at a time step~{\small\tt T}:
{\small \begin{verbatim}
goal(T) :- n {goal(I,T): agent(I)}, time(T).
\end{verbatim}}

\noindent The variation \gN\ of the gossip problem has negative goal conditions, which can be specified as in the example above.

\end{itemize}

Once the goal conditions are defined, we ensure that the plan reaches a goal state at some time step~{\small\tt T}:

{\small \begin{verbatim}
goal :- goal(T).
:- not goal.
\end{verbatim}}

\paragraph{Persistence of values of fluents}
Since the gossip problem does not involve forgetting, if an agent knows-whether a piece of information then he keeps that information:

{\small \begin{verbatim}
kww(I,info(K,D),T+1) :- kww(I,info(K,D),T), agent(I), info(K,D), time(T), T<m.
\end{verbatim}}

\vspace{-2ex}
\subsection{Actions: Calls between agents}

The output of the gossip problem is characterized by action constants -- atoms of the form {\small\tt call(i,j,t)} (``agent {\small \tt i} calls agent {\small \tt j} at time step {\small\tt t}.'')

\vspace{-2ex}
\paragraph{Action occurrences and preconditions}
Agent {\small\tt I} is free to call agent {\small\tt J} at any time, if he is allowed to (relative to the communication network):

{\small \begin{verbatim}
{call(I,J,T)} :- agent(I), agent(J), time(T), I!=J, connected(I,J), T<m.
\end{verbatim}}

\vspace{-2ex}
\paragraph{Effects of actions}
When an agent {\small\tt I} calls agent {\small\tt J} at time step {\small\tt T}, all permitted pieces of information {\small\tt K} of {\small\tt I} are passed to {\small\tt J}:

{\small \begin{verbatim}
kww(J,info(K,D),T+1) :- call(I,J,T), agent(I), agent(J),
  kww(I,info(K,D),T), info(K,D), permitted(I,J,K,T), time(T), T<m.
\end{verbatim}}

\noindent Furthermore, agent {\small\tt J} knows-whether agent {\small\tt I} knows-whether information {\small\tt K}:

{\small \begin{verbatim}
kww(J,info(kw(I,K),D+1),T+1) :- call(I,J,T), agent(I), agent(J), info(K,D), 
  kww(I,info(K,D),T), info(kw(I,K),D+1), permitted(I,J,K,T), time(T), T<m.
\end{verbatim}}

As in the original gossip problem, as an effect of {\small\tt call(I,J,T)}, all permitted pieces of information {\small\tt K} of {\small\tt J} are passed to {\small\tt I}:

{\small \begin{verbatim}
kww(I,info(K,D),T+1) :- call(I,J,T), agent(I), agent(J),
  kww(J,info(K,D),T), info(K,D), permitted(J,I,K,T), time(T), T<m.
kww(I,info(kw(J,K),D+1),T+1) :- call(I,J,T), agent(I), agent(J), info(K,D), 
  kww(J,info(K,D),T), info(kw(J,K),D+1), permitted(J,I,K,T), time(T), T<m.
\end{verbatim}}

\vspace{-2ex}
\paragraph{Permitted messages}
It can be assumed that all pieces of information are permitted to be shared at any time, unless told otherwise:

{\small \begin{verbatim}
permitted(I,J,K,T) :- connected(I,J), agent(I), agent(J), info(K,_), 
  kww(I,info(K,_),T), not -permitted(I,J,K,T), time(T), T<m.
\end{verbatim}}

\noindent Alternatively, agents can pass at least {\small\tt ll} and at most {\small\tt uu} pieces of information at any time:

{\small \begin{verbatim}
ll {permitted(I,J,K,T) : info(K,_), kww(I,info(K,_),T)} uu :-
  agent(I), agent(J), connected(I,J), time(T), T<m.
\end{verbatim}}

\vspace{-2ex}
\subsection{Constraints and preferences}

Concurrency can be prevented to some extent. For instance, an agent cannot be called while making a call, or by more than one agent at the same time. Also, an agent cannot call more than one agent at the same time. These concurrency constraints can be represented as follows:

{\small \begin{verbatim}
:- 2 {call(I,J,T): agent(J), connected(I,J);
  call(J1,I,T): agent(J1), connected(J1,I)}, agent(I), time(T), T<m.
\end{verbatim}}

\noindent Unless told otherwise, the formulation of calls as in the previous section, and with the constraint above,
allows parallel communication as in the variation \gC\ of the gossip problem.

We can also express preferences about occurrences of actions. For instance, the following ``weak'' constraint expresses our preference of less number of calls (as in the variation \gM\ of the gossip problem):

{\small \begin{verbatim}
:~ call(I,J,T), connected(I,J), agent(I;J), time(T), T<m. [3@1,I,J,T]
\end{verbatim}}

\noindent Intuitively, with these weak constraints, every atom of the form {\small\tt call(I,J,T)} included in an answer set~$S$ costs 3 units for $S$. Then the total cost of an answer set characterizes to what extent the preferences are not satisfied in a plan: the larger the total cost, the less preferred the plan is. The ASP solver \clingo\ finds an answer set whose total cost is minimum, and thus a plan with minimum number of calls.

The following weak constraint expresses our preference over the satisfaction of goal conditions at an earlier time step:

{\small \begin{verbatim}
:~ not goal(T), time(T). [3@2,T]
\end{verbatim}}

\noindent Intuitively, if a goal is not satisfied at time step {\small\tt T} in an answer set, then the total cost of the answer set is increased by 3. Therefore, with these weak constraints, the makespan of a plan is minimized. This optimization has a higher priority, i.e., of {\small\tt 2}, compared to the minimization of the number of calls. Therefore, \clingo\ first minimizes the makespan, and then tries to minimize the number of calls.

\vspace{-2ex}
\subsection{Epistemic planning problem}

For instance, for 4 agents, 4 messages, a complete communication network between agents, and an epistemic depth of 1, a solution to the gossip problem (with the conditions \gM, \gC) computed by the ASP solver~\clingo~\cite{GebserKKOSS11} for maximum time step 2 is as follows:

{\small \begin{verbatim}
call(1,2,0) call(4,3,0) call(2,3,1) call(1,4,1)
\end{verbatim}}

\noindent For epistemic depth of 2, the computed solution for maximum time step 4 is as follows:

{\small \begin{verbatim}
call(1,2,0) call(3,4,0) call(4,2,1) call(3,1,1)
call(4,1,2) call(4,2,3) call(1,3,3)
\end{verbatim}}


\vspace{-2ex}
\section{Evaluations}

We have performed some experiments for the gossip problem instances with the conditions \gM, \gC, where the goal is specified with respect to the epistemic depth 1 and the plan lengths are also minimized.  A complete communication network is considered between the agents. The maximum makespan is specified as~5.

In the experiments, we have used \clingo~5.2.2, and performed experiments on a Linux server with
3.30GHz Intel(R) Xeon(R) W-2155 CPU and 32 GB memory.
A maximum threshold of 10 seconds is provided to \clingo, allowing anytime search.

\begin{table}[htb]
\caption{Experimental results: For {\small\tt n} agents and the maximum time step {\small\tt m=5}, a solution to the gossip problem is found with anytime search of \clingo\ within 10 seconds. For each solution, whether it is optimal (O) or may be optimal (+), and the number of calls are reported.}
\label{tab:results-depth1}
\centering
\resizebox{\textwidth}{!}{\begin{tabular}{|c|c|c|c|c|c|c|c|c|c|c|c|c|c|c|}
\hline
{\small\tt n} & 2 & 3 & 4 & 5 & 6 & 7 & 8 & 9 & 10 & 11 & 12 & 13 & 14 & 15 \\
\hline
time (sec)& 0.004 & 0.005 & 0.005 & 0.082 & 0.049 & 10 & 3.685 & 10 &10 &10 &10 &10 &10 &10  \\
\hline
optimality  & O  & O  & O  & O  & O  & +  & O & +  & +  & +  & +  & +  & +  & +  \\
\hline
\# calls & 1 & 3 & 4 & 6 & 8 & 10 & 12 & 14 & 17 & 20 & 21 & 25 & 28 & 31\\
\hline
\end{tabular}}
\end{table}

\normalsize

The experimental results are presented in Table~\ref{tab:results-depth1}. As can be seen from the table, for {\small\tt n}=2,3,4,5,6,8, the solutions computed by \clingo are optimal with minimum number of calls; for {\small\tt n}=7 and {\small\tt n} larger than 8, the solutions computed by \clingo by anytime search may not be optimal.
If we increase the time threshold to 600 seconds, an optimal solution for {\small\tt n}=7 is computed in 336.928 seconds. For {\small\tt n}=11,13,15, slightly better solutions with 19, 24, 30 calls, respectively, are computed, but they may not be optimal.

\section{Discussions}
Gossip problems pose an interesting set of computational problems for Answer Set Programming (ASP). From the perspective of representation, it is challenging to model variations of gossip problems in an elaboration tolerant way and utilizing the useful constructs of ASP. From the perspective of computational performance, as can be seen from the preliminary experimental evaluations, although the ASP approach fares much better than the PDDL approach~\cite{CooperHMPV20}, it is still not scalable for larger instances. Our ongoing studies involve extension of experimental evaluations to other variations of gossip problems, and comparisons with the PDDL approach.

\bibliographystyle{eptcs}

\end{document}